\documentclass{article}

\usepackage[nonatbib,preprint]{neurips_2024}

\usepackage[utf8]{inputenc} 
\usepackage[T1]{fontenc}    
\usepackage{hyperref}       
\usepackage{url}            
\usepackage{booktabs}       
\usepackage{amsfonts}       
\usepackage{nicefrac}       
\usepackage{microtype}      
\usepackage{xcolor}         

\usepackage{multirow}
\usepackage{booktabs}
\usepackage{units}
\usepackage{pifont}

\usepackage{amsmath}
\usepackage{amssymb}
\usepackage{mathtools}
\usepackage{amsthm}
\usepackage{bm}

\usepackage{chngcntr}

\title{Hyper-Transformer for Amodal Completion}

\makeatletter
\def\thanks#1{\protected@xdef\@thanks{\@thanks
        \protect\footnotetext{#1}}}
\makeatother

\author{%
Jianxiong Gao$^{1*}$, Xuelin Qian$^{2*}$\thanks{*: Equal contribution.}, Longfei Liang$^{3}$, Junwei Han$^{2}$, Yanwei Fu$^{1\dagger}$ \thanks{$^\dagger$: Corresponding author} \\ 
$^1$Fudan University, $^2$Northwestern Polytechnical University \\
$^3$Shanghai Neuhelium Neuromorphic Intelligence Technology \\
\texttt{jxgao22@m.fudan.edu.cn, \{xlqian,jhan\}@nwpu.edu.cn} \\
\texttt{yanweifu@fudan.edu.cn}
}

\begin{document}

\maketitle

\begin{abstract}
Amodal object completion is a complex task that involves predicting the invisible parts of an object based on visible segments and background information. Learning shape priors is crucial for effective amodal completion, but traditional methods often rely on two-stage processes or additional information, leading to inefficiencies and potential error accumulation. To address these shortcomings, we introduce a novel framework named the \textbf{H}yper-\textbf{T}ransformer \textbf{A}modal \textbf{N}etwork (H-TAN). This framework utilizes a hyper transformer equipped with a dynamic convolution head to directly learn shape priors and accurately predict amodal masks. Specifically, H-TAN uses a dual-branch structure to extract multi-scale features from both images and masks. The multi-scale features from the image branch guide the hyper transformer in learning shape priors and in generating the weights for dynamic convolution tailored to each instance. The dynamic convolution head then uses the features from the mask branch to predict precise amodal masks. We extensively evaluate our model on three benchmark datasets: KINS, COCOA-cls, and D2SA, where H-TAN demonstrated superior performance compared to existing methods. Additional experiments validate the effectiveness and stability of the novel hyper transformer in our framework.
\end{abstract}

\section{Introduction}

Amodal segmentation, a critical task in computer vision, aims to identify the complete shapes of objects, even when parts are obscured by other objects. 
With segmentation models~\cite{kirillov2023segment, zhao2023fast, ke2024segment} growing in power,
attention has shifted from merely detecting visible parts to completing occluded portions, known as amodal completion.
This field holds particular importance in various practical applications, including autonomous driving~\cite{qian2023impdet,geiger2012we}, robotics~\cite{cheang2022learning,back2022unseen}, and augmented reality~\cite{park2008multiple,mathis2021fast}. For instance, in autonomous driving, a comprehensive view of partially visible objects is essential for safe navigation.

As we know, mainstream frameworks~\cite{li20222d,xiao2021amodal,ke2021deep,follmann2019learning,yao2022self,gao2023coarse} for amodal completion typically rely on shape priors. These methods often integrate additional information or adopt multitask learning approaches to comprehensively understand an object’s shape and predict its occluded parts. 
Techniques like Saovs~\cite{yao2022self}, which introduces optical flow, and A3D~\cite{li20222d} which incorporates 3D information, enhance the model's shape understanding but typically necessitate extra information, reducing convenience.
Methods such as VRSP~\cite{xiao2021amodal} and C2F-Seg~\cite{gao2023coarse} employ a two-stage training process: first learning a codebook to grasp shape priors and then reconstructing the object's shape. Although these methods achieve state-of-the-art performance, they suffer from computational inefficiency and potential error accumulation due to the two-stage training. Moreover, their fixed-parameter networks in the primary processing branches may struggle with diverse categories and complex occlusions, resulting in suboptimal results.

Drawing on the concept of dynamic convolution~\cite{chen2020dynamic}, we suppose a module that dynamically learns the unique characteristics of different objects while leveraging background information from images to discern intricate details about an object's shape priors. This approach facilitates the prediction of precise amodal masks for target objects. Naturally, this led us to employ a hypernetwork~\cite{ha2016hypernetworks} to learn a dynamic convolution head, essential for predicting amodal masks. Given the need to extract effective information from the entire image's background to learn shape priors, we integrated a transformer\cite{vaswani2017attention} with long-range perception capabilities into our hypernetwork.

\begin{figure}
\begin{center}
\includegraphics[width=\columnwidth]{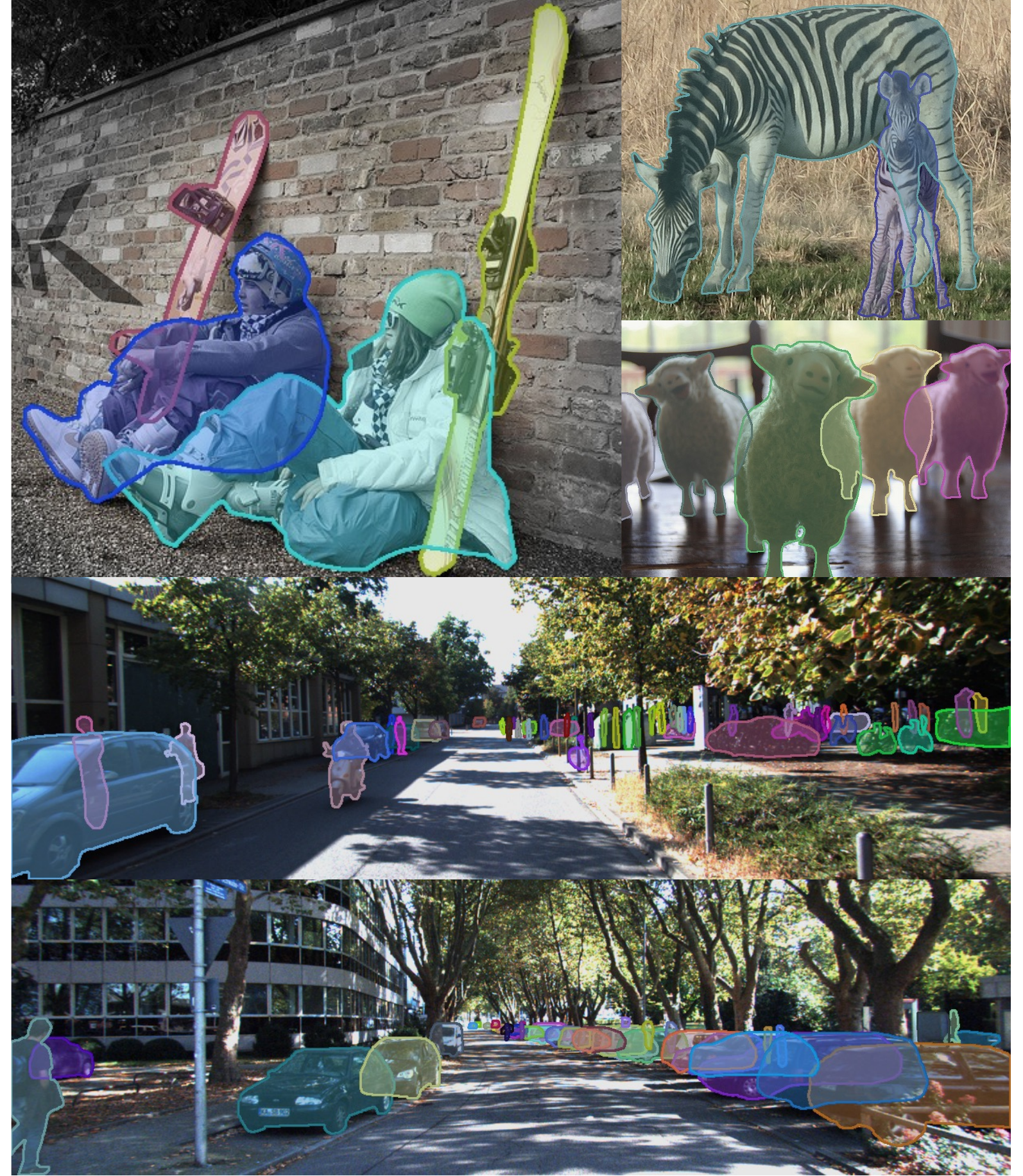}
\vskip -0.05in
\caption{\textbf{Visualization of Amodal Mask Quality Produced by Our Model.} We showcase the amodal masks of objects annotated in the KINS and COCOA-cls datasets.}
\label{fig:teaser}
\end{center}
\vskip -0.3in
\end{figure}

In this paper, we introduce a novel hyper transformer equipped with a dynamic convolution head. This design is tailored to learn the shape priors of each specific instance, where the features of the transformer are parameterized onto the dynamic convolution head, enhancing the accuracy of amodal mask predictions.

Our framework comprises two primary modules: a dual-branch structure and the hyper transformer with a dynamic convolution head. The dual-branch structure is divided into separate branches for images and masks. The image branch utilizes an adapted ResNet~\cite{he2016deep} to extract multi-scale features from RGB images, while the mask branch combines these features with mask details through skip connections, progressively refining the feature map of the amodal mask. We then integrate these multi-scale image features to guide the hyper transformer in generating specific weights for the kernels of the dynamic convolution head. These weights, along with the bias, enable the dynamic head to use the feature map from the mask branch to precisely predict the final amodal mask.

To assess the effectiveness of our model, we conducted tests on three well-known amodal benchmarks: KINS, COCOA-cls, and D2SA. Our model surpassed all previous benchmarks, establishing new performance records in this field.
Furthermore, we carried out detailed experiments on the hyper-parameters of the hyper transformer to elucidate our design choices and prove the potential of scaling up. Ablation studies were also performed to demonstrate the impact of each component within our model, validating their individual contributions to overall performance.

Our contributions can be summarized as follows:

\textbf{1)} We present the H-TAN, a novel framework for amodal segmentation. H-TAN incorporates an innovative hyper transformer designed to learn object shape priors and predict amodal masks using dynamic convolution.

\textbf{2)} Our model adopts a dual-branch approach for feature extraction. The image branch utilizes an adapted ResNet to extract multi-scale features from RGB images, while the mask branch integrates these features through skip connections to predict the feature map of the amodal masks.

\textbf{3)} Our experiments highlight the superiority of our method over other competitors, with H-TAN achieving new state-of-the-art performance.

\section{Related Work}

\subsection{Amodal Instance Segmentation.}
Amodal segmentation~\cite{zhu2017semantic} is a challenging task that involves predicting both the visible and occluded parts of objects. Numerous approaches~\cite{zhu2017semantic, follmann2019learning, xiao2021amodal, zhan2020self, tangemann2021unsupervised, ke2021occlusion, yang2019embodied, sun2022amodal} have sought to explore the occluded parts of objects using additional information or by learning the objects' shape priors, often achieving excellent performance. For example, Savos~\cite{yao2022self} employs optical flow, and A3D~\cite{li20222d} learns 3D shape priors to enhance performance. VRSP~\cite{xiao2021amodal} was the first to explicitly design a shape prior module with a codebook to refine the amodal mask. AISFormer~\cite{tran2022aisformer} enhances the extraction of long-range dependencies using transformers and employs multi-task training to develop a more comprehensive segmentation model. C2F-Seg~\cite{gao2023coarse} uses a masked transformer with a VQ codebook to generate a coarse mask, which is then refined with a CNN module. However, accessing additional information and the inefficiency of two-stage models, which can lead to error accumulation, are significant challenges.
In contrast, we propose a framework that does not require additional information and uses a single-stage learning process with a hyper transformer to learn the shape prior to each instance. This is followed by an instance-level dynamic head to predict the amodal mask, offering a streamlined and efficient alternative to existing methods.

\subsection{Hyper Network}  Hypernetworks~\cite{ha2016hypernetworks} pioneered the use of one network to generate the weights for another network, initially applying this innovative approach to image classification with notable success. Since then, the use of hypernetworks has expanded within the field of computer vision. This method has also achieved success in image manipulation~\cite{alaluf2022hyperstyle, muller2021overparametrization}. HyperSeg~\cite{nirkin2021hyperseg} utilizes a hypernetwork as an encoder to provide parameters for a decoder, achieving outstanding performance in semantic segmentation. CVAE-H~\cite{oh2022cvae} further extends the application of hypernetworks to the field of autonomous driving. Building on the successful experiences documented in previous literature, this paper explores the application of hypernetworks to amodal completion. Unlike previous approaches, we employ a transformer-based hypernetwork to learn the shape prior and generate a dynamic convolution head for each instance, offering a novel methodology in the process.

\section{Methodology}

\begin{figure}
\begin{center}
\includegraphics[width=\columnwidth]{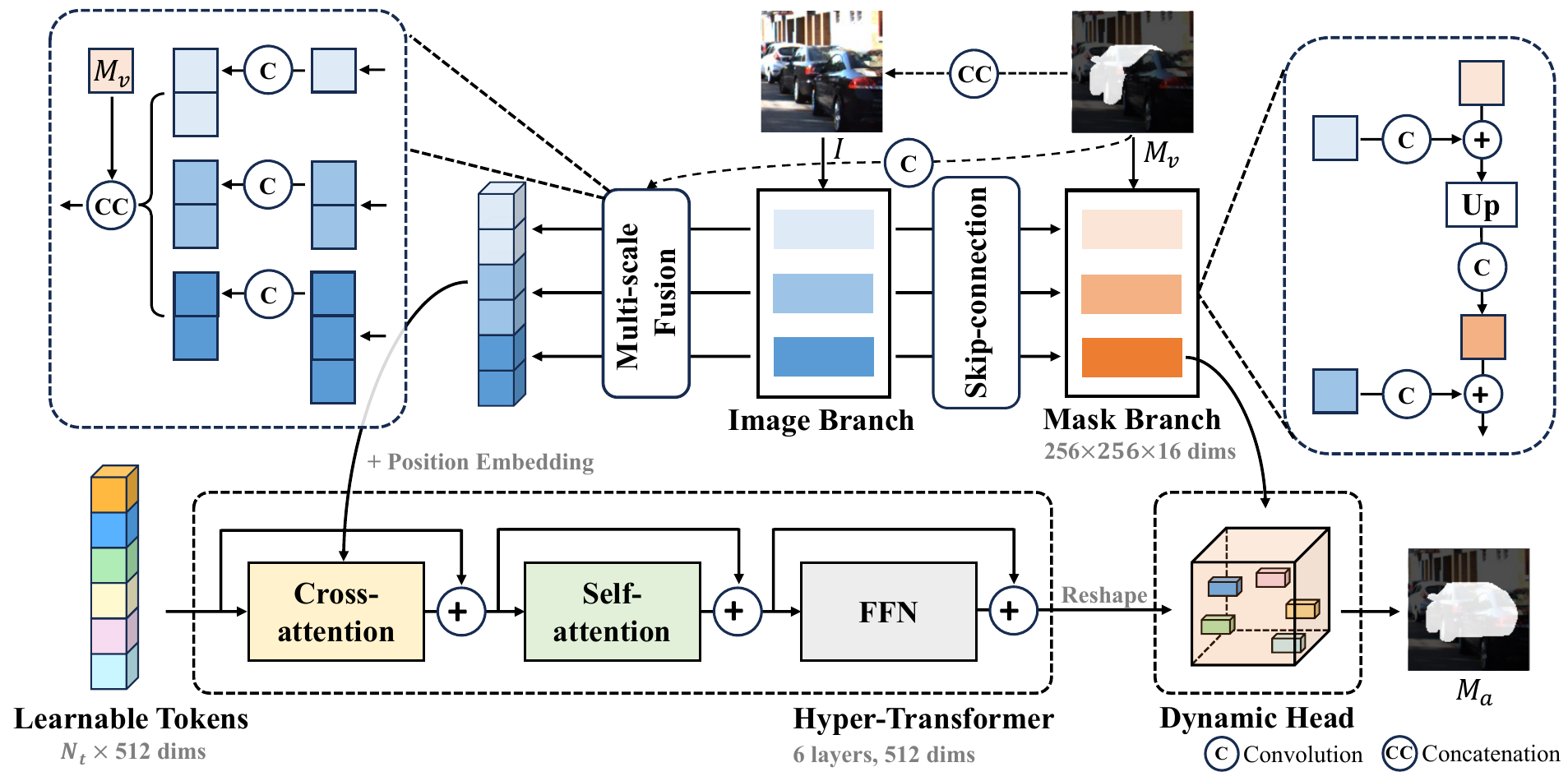}
\vskip -0.05in
\caption{Overview of the H-TAN framework. H-TAN extracts comprehensive features from both the image and mask through two branches. The image branch uses the mask to focus on the object that needs completion, extracting multi-scale features. The mask branch receives these multi-scale features from the image. Next, we design a Hyper Transformer and a Dynamic Head. The Hyper Transformer generates the parameters for the Dynamic Head guided by the image features, ultimately predicting the precise $M_a$.}
\label{fig:architecture}
\end{center}
\vskip -0.1in
\end{figure}

\textbf{Problem Setup}. Amodal object completion is designed to infer and reconstruct occluded parts of target objects in an image. The process involves taking an image $\bm{I}$ and a visible mask $\bm{M}_v$, which represents only the portions of the target objects visible in $\bm{I}$. The output is a comprehensive amodal mask $\bm{M}_a$, integrating both visible and occluded parts. Our objective is to synthesize $\bm{M}_a$ leveraging both $\bm{I}$ and $\bm{M}_v$. To achieve this, we introduce the \textbf{H}yper-\textbf{T}ransformer \textbf{A}modal \textbf{N}etwork (H-TAN), which consists of a dual-branch structure that separately extracts image and mask features, and a dynamic convolution head whose weights are generated by the hyper transformer.

\subsection{Dual Branches}

\noindent\textbf{Image Branch}.
The information from RGB images is critical for filling in the obscured parts of target objects. In the process of amodal completion, our focus is primarily on these occluded sections. Therefore, we naturally consider using the visible mask of the object to guide this branch. By directly locating the target object in the RGB image, this module focuses on the characteristics of the object itself as well as the background information of the image. Hence, we concatenate the image $\bm{I}$ and the visible mask $\bm{M}_v$ as the input of the branch:
\begin{equation}
\bm{X}_{input} = \text{concat}(\bm{I}, \bm{M}_v).
\end{equation}

To efficiently capture detailed features, we adopt a ResNet with a first layer 4-channel input as our encoder, which extracts multi-scale image features.

The input $\bm{X}_{input}$ passes through the ResNet layers, and we denote the feature maps in each layer as $I_{i}$, where $i$ represents the layer index. The output is described by:
\begin{equation}
I_{i} = \text{ResNetLayer}_i(\bm{X}_{input}).
\end{equation}

The four layers in ResNet capture different levels of detail and abstraction, and these features are crucial for the later stages of generating the weights for the dynamic convolution head and the completion of the amodal masks.

\textbf{Mask Branch}

The Mask Branch is designed to integrate the multi-scale features from the Image Branch with the visible mask $\bm{M}v$, to construct a detailed feature map for the amodal mask $\bm{M}a$. It progressively refines this feature map using the multi-scale features from the encoder. Initially, $\bm{M}v$ is resized to align with the spatial dimensions of $I{-1}$. Subsequent layers then sequentially merge these resized features with those from the Image Branch via skip connections. Each decoder layer enriches its output by incorporating corresponding encoder layer features, thereby enhancing detail in the progressively upsampled outputs. The upsampling process at each decoder layer is described by the equation:
\begin{equation}
\bm{D}i = \text{Up}(\bm{D}{i-1}) + C(I{i}).
\end{equation}
where $\bm{D}i$ represents the output from the $i$-th layer of the mask branch, $\text{Up}(\cdot)$ indicates the upsampling operation, and $C$ denotes the convolution layer that adjusts the feature $I{i}$ to match the dimensions of the mask feature. The final decoder output, $\bm{D}{final}$, is fed into the dynamic convolution head. The head's dynamically generated kernel weights, derived from the hyper transformer, are used to accurately predict the amodal mask $\bm{M}_a$.

\subsection{Hyper Transformer}

Understanding the shape priors of objects that are partially occluded is key to accurately completing their full shapes, which can vary widely from one object to another. To effectively learn these shape priors, we developed a novel hyper transformer. This transformer is designed to capture the distinctive characteristics of each object and make full use of the background information and the relationships with objects that obscure them. It then generates the weights for the dynamic convolution head. This method allows for a customized convolution head for each instance, tailored to the specific needs of that object's shape completion. Specifically, we fuse the multi-scale features from the image branch and use these fused features to guide the hyper transformer to generate the weights for the convolution layers in the dynamic head.

In more detail, each feature map $I_i$ from the encoder's ResNet layers is processed through a convolution layer $c_i(\cdot)$, treating each separately. These adapted features are then combined along the feature dimension to create a comprehensive multi-scale feature representation $I_f$:
\begin{equation}
I_f = \text{concat}(c_1(I_1), c_2(I_2), c_3(I_3), c_4(I_4)).
\end{equation}

We also process the visible mask $\bm{M}_v$ to derive mask embeddings $E_{\text{mask}}$, which are concatenated with $I_f$ to enhance the representation. Cosine-sine positional embeddings are added to improve spatial awareness, resulting in a fully integrated feature set used by the hyper transformer:
\begin{equation}
F_c = \text{concat}(I_f + \text{pos\_embedding}, E_{\text{mask}} + \text{pos\_embedding}).
\end{equation}

Next, we initialize a learnable token $E_l$ with shape $(B, N_t, D)$, where $B$ is the batch size, $D$ is the hidden dimension of the Hyper Transformer, and $N_t$ is the total number of tokens we need to learn. The calculation of $N_t$ will be discussed in Sec.~\ref{dynamichead}.

Within the hyper transformer, the integrated feature $F_c$ guides the learnable token $E_l$ through cross-attention, emphasizing feature relevance for better processing to get $E_l^{\text{cross}}$. This is followed by self-attention on $E_l^{\text{cross}}$, and then the output $E_l^{\text{self}}$ is processed through a feedforward network (FFN):
\begin{equation}
E_l^{\text{cross}} = \text{CrossAttention}(E_l, F_c)
\end{equation}
\begin{equation}
E_l^{\text{self}} = \text{SelfAttention}(E_l^{\text{cross}})
\end{equation}
\begin{equation}
E_l = \text{FFN}(E_l^{\text{self}})
\end{equation}
After passing through $L$ transformer blocks, the learnable token $E_l$ parameterizes the dynamic head. The detailed hyper-parameters of the hyper transformer will be discussed in Sec.~\ref{Discussion} in detail to illustrate how they affect performance.

\subsection{Dynamic Head}
\label{dynamichead}

The dynamic head is designed to predict the amodal mask $\bm{M}_a$ based on instance-specific weights, essentially capturing the shape prior. We specify the size of the dynamic head and the number $N_t$ of tokens needed to learn in the hyper transformer.

We calculate the number of the tokens needed for the convolution layers in the Dynamic Head as follows:
\begin{equation}
N_t = (C^1_{\text{in}} \cdot K_s^1)^2 + C^1_{\text{out}}+(C^2_{\text{in}} \cdot K_s^2)^2 + C^2_{\text{out}}+(C^3_{\text{in}} \cdot K_s^3)^2 + C^3_{\text{out}},
\end{equation}
where $N_t$ represents the total number of parameters, $C^i_{\text{in}}$ and $C^i_{\text{out}}$ are the numbers of input and output channels of the $i$th convolution layer, and $K_s^i$ is the kernel size. 

\begin{figure}
\begin{center}
\includegraphics[width=\columnwidth]{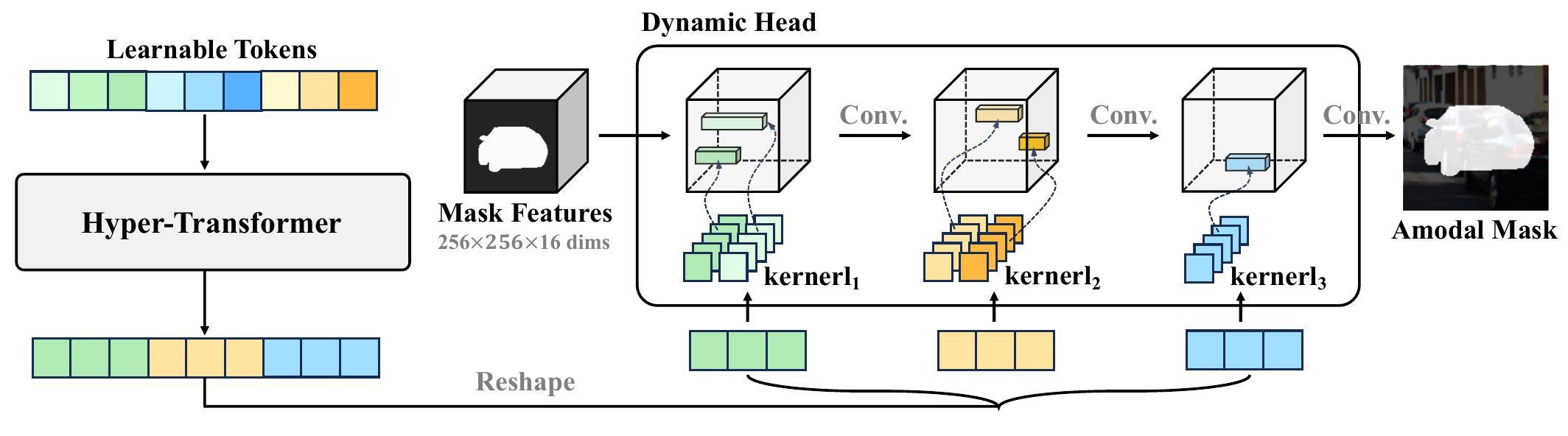}
\vskip -0.05in
\caption{Details of the use of learnable tokens in the hyper transformer as weights for the dynamic head, enabling precise prediction of the amodal mask based on mask features.}
\label{fig:weights}
\end{center}
\vskip -0.1in
\end{figure}

In practice, the process of reshaping the features of the hyper transformer and then treating them as the weights and biases for the convolution kernels in the dynamic head are shown in Fig.~\ref{fig:weights}. The learnable tokens $E_l$ are then utilized by the dynamic head to predict the target amodal mask $\bm{M}_a$ based on the features extracted by the mask branch.
\begin{equation}
\bm{M}_a = \text{Dynamic Head}(D_{-1}).
\end{equation}

Our model's training objective is to minimize the binary cross-entropy loss between the predicted and actual amodal masks:
\begin{equation}
\mathcal{L} \coloneqq -\frac{1}{N} \sum_{i=1}^N \left[ \bm{M}_a^{(i)} \log(\hat{\bm{M}}_a^{(i)}) + (1 - \bm{M}_a^{(i)}) \log(1 - \hat{\bm{M}}_a^{(i)}) \right]
\end{equation}

During training, the hyper transformer learns the features in batches, while in the dynamic head, weights are learned for each instance. Thus, we predict the amodal mask $\bm{M}_a$ at the instance level. This process remains consistent during the inference step as well. This structured approach allows our model to efficiently learn and accurately predict the complete shapes of occluded objects, enhancing the precision of amodal completion.


\section{Experiments}

\subsection{Experiments Details}

\noindent \textbf{Datasets}
To evaluate the efficacy of our proposed model, we conduct comprehensive experiments on both image and video amodal segmentation benchmarks. 
\textbf{1) KINS}~\cite {qi2019amodal} is a large-scale amodal instance dataset, which is built upon KITTI \cite{geiger2012we}. It contains 7 categories that are common on the road, including car, truck, pedestrian, \textit{etc}. There are 14,991 manually annotated images in total, 7,474 of which are used for training and the remaining for testing.
\textbf{2) COCOA-cls}~\cite{ehsani2018segan} is derived from COCO dataset~\cite{lin2014microsoft}. It consists of 2,476 images in the training set and 1,223 images in the testing set. There are 80 objects in this dataset.
\textbf{3) D2SA}~\cite{follmann2019learning} is generated from the D2A dataset~\cite{follmann2018mvtec} by overlapping instances. D2SA comprises 60 categories of instances, including 2,000 images in the training set and 3,600 images in the validation set. 

\noindent \textbf{Metric}
In terms of metric, we use standard metric prevalent in many amodal completion literature~\cite{li20222d, li2023gin, gao2023coarse, yao2022self}. Specifically, we utilize mean Intersection over Union (mean-IoU) to evaluate the quality of the predicted amodal masks. Mean-IoU consists of two components: mIoU${\text{full}}$ and mIoU${\text{occ}}$. mIoU${\text{full}}$ is calculated by comparing the predicted masks with the GT amodal mask, whereas mIoU${\text{occ}}$ specifically assesses the occluded regions. These metrics directly gauge the accuracy of amodal completion. Importantly, mIoU$_{\text{occ}}$ is crucial for assessing the effectiveness of amodal completion, as it quantifies how precisely the masks cover the occluded parts of the target objects.

\noindent \textbf{Implementation Details}.
\label{Implementation}
Our framework is implemented on the PyTorch platform. For a fair comparison, we use the pre-trained ResNet-50, except for the first convolution layer, to initialize the image branch, while the other modules of H-TAN are trained from scratch. In terms of architecture design, we set the number of transformer layers to 6, with a feature dimension of 512 and 8 heads. The number of learnable tokens in the hyper transformer is set to $N_t = 153$, and the number of input channels for the Dynamic Head is $C_{in} = 8$. For data augmentation, morphology dilation, erosion, and Gaussian blur are applied to the visible mask $M_v$ inputs. We adopt the AdamW optimizer~\cite{loshchilov2017decoupled} with a learning rate of 3e-4 and a warm-up learning schedule for all experiments. The model is trained on three datasets using a batch size of 48 on 24GB RTX 3090 GPUs. The total number of training iterations is set to 40K for KINS, 8K for COCOA, and 10K for D2SA. To demonstrate the efficacy of our model, we compare it with several SOTA methods: Mask-RCNN~\cite{he2017mask}, PCNET~\cite{zhan2020self}, VRSP~\cite{xiao2021amodal}, A3D~\cite{li20222d}, AISformer~\cite{tran2022aisformer}, GIN~\cite{li2023gin}, and C2F-Seg~\cite{gao2023coarse}.

\begin{figure}
\begin{center}
\includegraphics[width=\columnwidth]{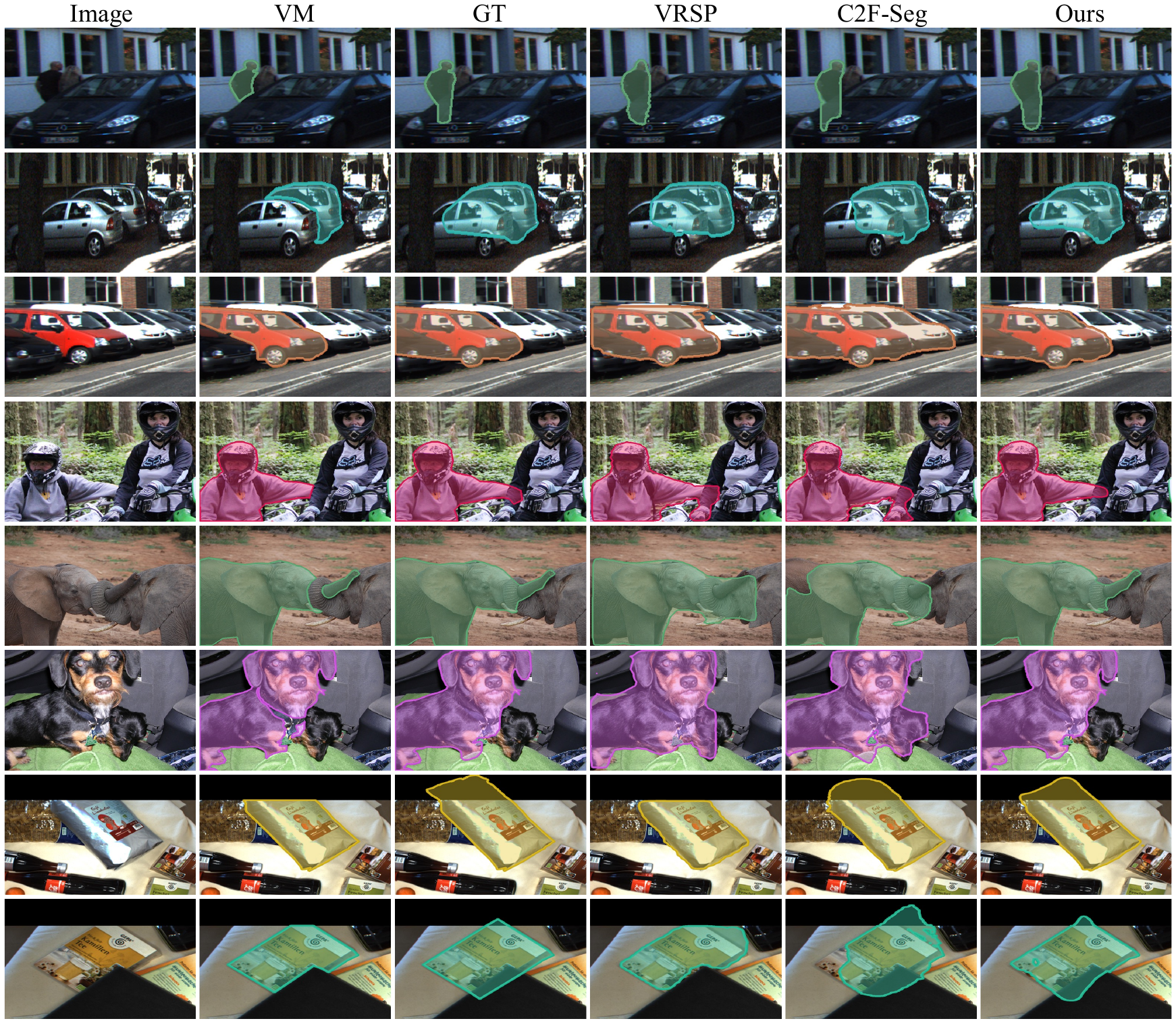}
\vskip -0.05in
\caption{Qualitative results compared with VRSP, C2F-Seg, and our H-TAN on KINS, COCOA-cls, and D2SA datasets. VM and GT denote the ground-truth visible mask and amodal mask, respectively. Best viewed in color and zoomed in for details.}
\label{fig:visualization}
\end{center}
\vskip -0.1in
\end{figure}

\begin{table} 
\centering
\setlength{\tabcolsep}{1mm}{
\caption{
\textbf{Quantitative results on KINS, COCOA-cls and D2SA.} We report and compare the Mean-IoU metrics for KINS, COCOA-cls and D2SA of our model with baselines.
\label{tab:metrics}
}
\vskip -0.05in
{\begin{tabular}{lccccccc}
\toprule
\multicolumn{1}{c}{\multirow{2}{*}{\textsc{Methods}}} & \multicolumn{1}{c}{\multirow{2}{*}{\textsc{Shape Prior}}} & \multicolumn{2}{c}{\textbf{KINS}} & \multicolumn{2}{c}{\textbf{COCOA-cls}} & \multicolumn{2}{c}{\textbf{D2SA}}  \\
\cmidrule{3-8}
 &  & mIoU$_{full}$ & mIoU$_{occ}$ & mIoU$_{full}$ & mIoU$_{occ}$  & mIoU$_{full}$ & mIoU$_{occ}$  
 \\
\midrule
\midrule
Mask-RCNN~\cite{he2017mask}       & \ding{56}  & 60.13 & -     & 63.85  & -     & 74.62  & - \\
A3D~\cite{li20222d}               & 3D  & 61.43 & -     & 64.27  & -     & 74.71  & - \\
GIN~\cite{li2023gin}              & ShapeDict  & 68.31 & -     & 72.46  & -     & 78.23  & - \\
PCNET~\cite{zhan2020self}         & \ding{56}  & 78.02 & 38.14 & 76.91  & 20.34 & 80.45  & 28.56 \\
AISformer~\cite{tran2022aisformer}& \ding{56} & 81.53 & 48.54 & 72.69  & 13.75 & 86.81  & 30.01 \\
VRSP~\cite{xiao2021amodal}        & Codebook  & 80.70 & 47.33 & 78.98  & 22.92 & 88.08  & 35.17 \\
C2F-Seg~\cite{gao2023coarse}      & Codebook & 87.89 & 57.60 & 87.13  & 36.55 & 91.04  & 42.72 \\
\midrule
Ours  & Hyper   & \textbf{91.55} & \textbf{69.12} & \textbf{92.37} & \textbf{49.24} & \textbf{94.71}  & \textbf{58.17} \\
\bottomrule
\end{tabular}}}
\vskip -0.05in
\end{table}

\subsection{Experiments Results}
We compare the performance of our H-TAN with several state-of-the-art (SOTA) methods on three common amodal benchmarks to showcase our model's performance. As shown in Tab.~\ref{tab:metrics}, we present a comparison of our model against several baseline methods across three different datasets: KINS, COCOA-cls, and D2SA. Our model achieves new SOTA results on all datasets, particularly outperforming C2F-Seg by a large margin in mIoU${\text{occ}}$, with improvements of at least 10 points. This significant improvement in mIoU${\text{occ}}$ demonstrates our model's superior ability to handle occlusions, a common challenge in object segmentation tasks.
We also include qualitative results comparing our model with VRSP and C2F-Seg to further illustrate these advancements, as shown in Fig.~\ref{fig:visualization}. VRSP~\cite{xiao2021amodal} and C2F-Seg~\cite{gao2023coarse} struggle to accurately differentiate between the target object and the occluding object when dealing with the same category, resulting in poor performance at the boundaries and occluded parts, as seen in examples 3, 4, 5, and 6. In contrast, H-TAN learns the relationship between the target object and surrounding objects at the instance level, allowing the dynamic head to predict accurate amodal masks. Additionally, our model more accurately estimates the depth of occlusion for target objects, such as people and cars, which is highly beneficial in autonomous driving scenarios.
Overall, our experiments demonstrate that our method sets a new SOTA on real-world datasets, establishing a new benchmark for future research in the field.

\begin{table}
\centering
\setlength{\tabcolsep}{3mm}{
\caption{
\textbf{Exploration of Transformer Scale.} We report mIoU${full}$ and mIoU${occ}$ on KINS and COCOA-cls with $C_{in} = 8$ to test the performance of H-TAN as the hyper transformer scales up.
\label{tab:ablation_scale_transformer}
}
\vskip -0.05in
{\begin{tabular}{ccccccc}
\toprule
\multicolumn{1}{c}{\multirow{2}{*}{\textsc{Depth}}} & \multicolumn{1}{c}{\multirow{2}{*}{\textsc{Head}}} &  \multicolumn{1}{c}{\multirow{2}{*}{\textsc{Dimension}}} & \multicolumn{2}{c}{\textbf{KINS}} & \multicolumn{2}{c}{\textbf{COCOA-cls}} \\
\cmidrule{4-7}
& &  & mIoU$_{full}$ & mIoU$_{occ}$  & mIoU$_{full}$ & mIoU$_{occ}$ \\
\midrule
\midrule
6  & 8  & 256 & 91.42 & 68.09     & 92.15  & 48.02       \\
6  & 8  & 512 & 91.55 & 69.12     & 92.37  & 49.24       \\ 
\midrule
8  & 8  & 512 & 91.70 & 69.92     & 92.44  & 50.01       \\
8  & 12 & 768 & \textbf{91.84} & 70.32     & \textbf{92.48}  & \textbf{50.29}       \\
8  & 16 & 1024 & 91.82 & \textbf{70.46}     & 92.41  & 50.17       \\
\midrule
12 & 16 & 1024 & 91.73 & 70.01     & 92.25  & 48.34     \\
\bottomrule
\end{tabular}}}
\vskip -0.1in
\end{table}

\subsection{Discussion on the Hyper Transformer}
\label{Discussion}
As for such a novel Hyper Transformer, we discuss and conduct experiment to show how us select the hyper-parameter for it and explore the robustness in amodal segmentation. To be consistent, the following experiments are all on KINS and COCOA-cls.

\textbf{1) Scale of Transformer.} First, we conduct experiments to explore the impact of the transformer's scale on our model's performance. We train the model with six different configurations of depth, feature dimension, and head count. To ensure fairness, we fix the number of input channels for the dynamic head at $C_{in} = 8$. As detailed in Tab.~\ref{tab:ablation_scale_transformer}, we report both the mIoU${full}$ and mIoU${occ}$. The results indicate that our hyper transformer can indeed improve performance when scaled up appropriately.

\textbf{2) Effect of $C_{in}$ in Dynamic Head.} Besides the effect of the hyper-transformer, we also need to evaluate the performance of our model with different values of $C_{in}$, i.e., explore the effect of the number of learnable tokens $N_t$ on the performance of our model. Experiments were performed with $C_{in}$ values of 2, 4, 8, and 16. Correspondingly, the number of learnable tokens $N_t$ in the hyper transformer was 13, 45, 153, and 561, respectively. To ensure fairness, we fixed the hyper transformer size to a depth of 6, a feature dimension of 512, and 8 heads. Results in Tab.~\ref{tab:ablation_inch} show that increasing $C_{in}$ improves the performance of H-TAN on large-scale datasets like KINS, while it slightly decreases performance on COCOA-cls due to the smaller size of this dataset.

Overall, the two experiments demonstrate the potential and stability of the hyper transformer with the dynamic head when scaled up on large datasets, as KINS is a little larger than COCOA-cls.

\begin{table}
\centering
\setlength{\tabcolsep}{6mm}{
\caption{
\textbf{Exploration of $C_{in}$ in Dynamic Head.} We report mIoU$_{full}$ and mIoU$_{occ}$ on KINS and COCOA-cls with a fixed hyper transformer having 6 layers and a 512 feature dimension to test the performance of H-TAN as $C_{in}$ increases.
\label{tab:ablation_inch}
}
\vskip -0.05in
{\begin{tabular}{lccccc}
\toprule
\multicolumn{1}{c}{\multirow{2}{*}{$N$}} & \multicolumn{1}{c}{\multirow{2}{*}{C}} & \multicolumn{2}{c}{\textbf{KINS}} & \multicolumn{2}{c}{\textbf{COCOA-cls}} \\
\cmidrule{3-6}
 & & mIoU$_{full}$ & mIoU$_{occ}$ & mIoU$_{full}$ & mIoU$_{occ}$ \\
\midrule
\midrule
25    &  2  & 91.35 & 67.87     & 91.82  & 91.82    \\
98    &  4  & 91.38 & 67.58     & 92.03  & 45.39    \\
193   &  8  & \textbf{91.55} & 69.12     & \textbf{92.24}  & \textbf{47.79}    \\
562   &  16 & 91.53 & \textbf{69.34}     & 91.72  & 46.67     \\
\bottomrule
\end{tabular}}}
\vskip -0.1in
\end{table}

\subsection{Ablation Study}
For our proposed framework H-TAN, we conducted several ablation studies to evaluate the necessity and impact of the components on performance.

\noindent\textbf{Ablation for the component in H-TAN}. 
\textbf{1)} The most crucial ablation study is to assess whether the hyper transformer combined with the dynamic head contributes positively to our model's performance. We attempt to remove this module, retaining only the dual branches to predict the amodal mask.
\textbf{2)} We remove the multi-scale fusion module that follows the image branch and instead input only the features from the last layer of the image branch to the hyper transformer.
\textbf{3)} We remove the skip connections between the image branch and the mask branch.
We present the results of several ablation studies and compare them with C2F-Seg~\cite{gao2023coarse} in Tab.~\ref{tab:ablation_hyper}. The experiments demonstrate the effectiveness of each module, especially the hyper transformer and the multi-scale fusion. Additionally, the experiments show that better image features help the model to learn the background information of each object, which in turn improves the learning of shape priors for predicting amodal masks.

\begin{table}
\centering
\setlength{\tabcolsep}{4.1mm}{
\caption{
\textbf{Ablation study for components in H-TAN.} We report mIoU$_{full}$ and mIoU$_{occ}$ on KINS and COCOA-cls to evaluate the impact of the hyper transformer. 
\label{tab:ablation_hyper}
}
\vskip -0.05in
{\begin{tabular}{lcccc}
\toprule
\multicolumn{1}{c}{\multirow{2}{*}{\textsc{Methods}}}  & \multicolumn{2}{c}{\textbf{KINS}} & \multicolumn{2}{c}{\textbf{COCOA-cls}} \\
\cmidrule{2-5}
  & mIoU$_{full}$ & mIoU$_{occ}$ & mIoU$_{full}$ & mIoU$_{occ}$ \\
\midrule
\midrule
C2F-Seg                      & 87.89 & 57.60     & 87.13  & 36.55    \\
\midrule
H-TAN (Ours)                   & \textbf{91.55} & \textbf{69.12}  & \textbf{92.24}  & \textbf{47.79}    \\
\quad \textit{w/o} Hyper Transformer        & 85.98 & 56.15     & 86.17  & 33.56    \\
\quad \textit{w/o} Multi-scale Fusion       & 89.68 & 60.47     & 90.96  & 44.07    \\
\quad \textit{w/o} Skip Connection          & 91.01 & 65.96     & 91.31  & 45.33    \\
\bottomrule
\end{tabular}}}
\vskip -0.1in
\end{table}

\begin{table}
\centering
\setlength{\tabcolsep}{3mm}{
\caption{
\textbf{Ablation Study for $\bm{M}_v$ in Each Component.} We report and compare mIoU$_{full}$ and mIoU$_{occ}$ to show whether the input of the visible mask $\bm{M}_v$ improves the performance of our model.
\label{tab:ablation_vm}
}
\vskip -0.07in
{\begin{tabular}{ccccccc}
\toprule
\multicolumn{1}{c}{\multirow{2}{*}{\textsc{Image}}} & \multicolumn{1}{c}{\multirow{2}{*}{\textsc{Mask}}} &  \multicolumn{1}{c}{\multirow{2}{*}{\textsc{Dynamic Head}}} & \multicolumn{2}{c}{\textbf{KINS}} & \multicolumn{2}{c}{\textbf{COCOA-cls}} \\
\cmidrule{4-7}
& &  & mIoU$_{full}$ & mIoU$_{occ}$  & mIoU$_{full}$ & mIoU$_{occ}$ \\
\midrule
\midrule
\ding{55}   &  \ding{51}  & \ding{51} & 90.11 & 64.53     & 90.15  & 43.51       \\
\ding{51}   &  \ding{55}  & \ding{51} & 91.27 & 67.29     & 91.21  & 45.06       \\
\ding{51}   &  \ding{51}  & \ding{55} & 91.40 & 68.76     & 91.49  & 45.87       \\
\ding{51}   &  \ding{51}  & \ding{51} & 91.55 & 69.12     & 92.24  & 47.79       \\
\bottomrule
\end{tabular}}}
\vskip -0.1in
\end{table}

\noindent\textbf{Ablation for $\bm{M}_v$ in each component}. 
\label{Ablationmv}
In our model, we utilize the visible mask $\bm{M}_v$ in each module for distinct purposes. We conducted experiments to validate the effectiveness of integrating $\bm{M}_v$ within the image branch, mask branch, and dynamic convolution head. These tests were performed on the KINS and COCOA-cls datasets, with results detailed in Tab.~\ref{tab:ablation_vm}. The findings suggest that incorporating $\bm{M}_v$ into various modules yields slight improvements in our model’s performance. Notably, the presence of $\bm{M}_v$ in the image branch is crucial, indicating its significant role in helping to pinpoint target objects and focus more on occluded parts.

\section{Conclusion}
In this work, we introduce a novel framework, H-TAN, for amodal completion, which utilizes a dual branch structure to extract features from images and masks. Based on these features, it employs an innovative hyper transformer to generate weights for a dynamic convolution head tailored to each target object, enabling precise prediction of amodal masks. We conducted extensive experiments to demonstrate the stability of the hyper transformer and its potential for scaling up. Additionally, through ablation studies, we validated the effectiveness of each component of our model. The experimental results confirm that our approach, leveraging a hypernetwork, effectively learns shape priors from background information, leading to precise amodal masks and achieving new state-of-the-art performance in image amodal completion.


{\small
\bibliographystyle{plain}
\bibliography{egbib}

\begin{thebibliography}{10}

\bibitem{alaluf2022hyperstyle}
Yuval Alaluf, Omer Tov, Ron Mokady, Rinon Gal, and Amit Bermano.
\newblock Hyperstyle: Stylegan inversion with hypernetworks for real image editing.
\newblock In {\em Proceedings of the IEEE/CVF conference on computer Vision and pattern recognition}, pages 18511--18521, 2022.

\bibitem{back2022unseen}
Seunghyeok Back, Joosoon Lee, Taewon Kim, Sangjun Noh, Raeyoung Kang, Seongho Bak, and Kyoobin Lee.
\newblock Unseen object amodal instance segmentation via hierarchical occlusion modeling.
\newblock In {\em 2022 International Conference on Robotics and Automation (ICRA)}, pages 5085--5092. IEEE, 2022.

\bibitem{cheang2022learning}
Chilam Cheang, Haitao Lin, Yanwei Fu, and Xiangyang Xue.
\newblock Learning 6-dof object poses to grasp category-level objects by language instructions.
\newblock In {\em 2022 International Conference on Robotics and Automation (ICRA)}, pages 8476--8482. IEEE, 2022.

\bibitem{chen2020dynamic}
Yinpeng Chen, Xiyang Dai, Mengchen Liu, Dongdong Chen, Lu~Yuan, and Zicheng Liu.
\newblock Dynamic convolution: Attention over convolution kernels.
\newblock In {\em Proceedings of the IEEE/CVF conference on computer vision and pattern recognition}, pages 11030--11039, 2020.

\bibitem{ehsani2018segan}
Kiana Ehsani, Roozbeh Mottaghi, and Ali Farhadi.
\newblock Segan: Segmenting and generating the invisible.
\newblock In {\em Proceedings of the IEEE conference on computer vision and pattern recognition}, pages 6144--6153, 2018.

\bibitem{follmann2018mvtec}
Patrick Follmann, Tobias Bottger, Philipp Hartinger, Rebecca Konig, and Markus Ulrich.
\newblock Mvtec d2s: densely segmented supermarket dataset.
\newblock In {\em Proceedings of the European conference on computer vision (ECCV)}, pages 569--585, 2018.

\bibitem{follmann2019learning}
Patrick Follmann, Rebecca K{\"o}nig, Philipp H{\"a}rtinger, Michael Klostermann, and Tobias B{\"o}ttger.
\newblock Learning to see the invisible: End-to-end trainable amodal instance segmentation.
\newblock In {\em 2019 IEEE Winter Conference on Applications of Computer Vision (WACV)}, pages 1328--1336. IEEE, 2019.

\bibitem{gao2023coarse}
Jianxiong Gao, Xuelin Qian, Yikai Wang, Tianjun Xiao, Tong He, Zheng Zhang, and Yanwei Fu.
\newblock Coarse-to-fine amodal segmentation with shape prior.
\newblock In {\em Proceedings of the IEEE/CVF International Conference on Computer Vision}, pages 1262--1271, 2023.

\bibitem{geiger2012we}
Andreas Geiger, Philip Lenz, and Raquel Urtasun.
\newblock Are we ready for autonomous driving? the kitti vision benchmark suite.
\newblock In {\em 2012 IEEE conference on computer vision and pattern recognition}, pages 3354--3361. IEEE, 2012.

\bibitem{ha2016hypernetworks}
David Ha, Andrew Dai, and Quoc~V Le.
\newblock Hypernetworks.
\newblock {\em arXiv preprint arXiv:1609.09106}, 2016.

\bibitem{he2017mask}
Kaiming He, Georgia Gkioxari, Piotr Doll{\'a}r, and Ross Girshick.
\newblock Mask r-cnn.
\newblock In {\em Proceedings of the IEEE international conference on computer vision}, pages 2961--2969, 2017.

\bibitem{he2016deep}
Kaiming He, Xiangyu Zhang, Shaoqing Ren, and Jian Sun.
\newblock Deep residual learning for image recognition.
\newblock In {\em Proceedings of the IEEE conference on computer vision and pattern recognition}, pages 770--778, 2016.

\bibitem{ke2021deep}
Lei Ke, Yu-Wing Tai, and Chi-Keung Tang.
\newblock Deep occlusion-aware instance segmentation with overlapping bilayers.
\newblock In {\em Proceedings of the IEEE/CVF conference on computer vision and pattern recognition}, pages 4019--4028, 2021.

\bibitem{ke2021occlusion}
Lei Ke, Yu-Wing Tai, and Chi-Keung Tang.
\newblock Occlusion-aware video object inpainting.
\newblock In {\em Proceedings of the IEEE/CVF International Conference on Computer Vision}, pages 14468--14478, 2021.

\bibitem{ke2024segment}
Lei Ke, Mingqiao Ye, Martin Danelljan, Yu-Wing Tai, Chi-Keung Tang, Fisher Yu, et~al.
\newblock Segment anything in high quality.
\newblock {\em Advances in Neural Information Processing Systems}, 36, 2024.

\bibitem{kirillov2023segment}
Alexander Kirillov, Eric Mintun, Nikhila Ravi, Hanzi Mao, Chloe Rolland, Laura Gustafson, Tete Xiao, Spencer Whitehead, Alexander~C Berg, Wan-Yen Lo, et~al.
\newblock Segment anything.
\newblock In {\em Proceedings of the IEEE/CVF International Conference on Computer Vision}, pages 4015--4026, 2023.

\bibitem{li20222d}
Zhixuan Li, Weining Ye, Tingting Jiang, and Tiejun Huang.
\newblock 2d amodal instance segmentation guided by 3d shape prior.
\newblock In {\em European Conference on Computer Vision}, pages 165--181. Springer, 2022.

\bibitem{li2023gin}
Zhixuan Li, Weining Ye, Tingting Jiang, and Tiejun Huang.
\newblock Gin: Generative invariant shape prior for amodal instance segmentation.
\newblock {\em IEEE Transactions on Multimedia}, 2023.

\bibitem{lin2014microsoft}
Tsung-Yi Lin, Michael Maire, Serge Belongie, James Hays, Pietro Perona, Deva Ramanan, Piotr Doll{\'a}r, and C~Lawrence Zitnick.
\newblock Microsoft coco: Common objects in context.
\newblock In {\em European conference on computer vision}, pages 740--755. Springer, 2014.

\bibitem{loshchilov2017decoupled}
Ilya Loshchilov and Frank Hutter.
\newblock Decoupled weight decay regularization.
\newblock {\em arXiv preprint arXiv:1711.05101}, 2017.

\bibitem{mathis2021fast}
Florian Mathis, John~H Williamson, Kami Vaniea, and Mohamed Khamis.
\newblock Fast and secure authentication in virtual reality using coordinated 3d manipulation and pointing.
\newblock {\em ACM Transactions on Computer-Human Interaction (ToCHI)}, 28(1):1--44, 2021.

\bibitem{muller2021overparametrization}
Lorenz~K Muller.
\newblock Overparametrization of hypernetworks at fixed flop-count enables fast neural image enhancement.
\newblock In {\em Proceedings of the IEEE/CVF Conference on Computer Vision and Pattern Recognition}, pages 284--293, 2021.

\bibitem{nirkin2021hyperseg}
Yuval Nirkin, Lior Wolf, and Tal Hassner.
\newblock Hyperseg: Patch-wise hypernetwork for real-time semantic segmentation.
\newblock In {\em Proceedings of the IEEE/CVF Conference on Computer Vision and Pattern Recognition}, pages 4061--4070, 2021.

\bibitem{oh2022cvae}
Geunseob Oh and Huei Peng.
\newblock Cvae-h: Conditionalizing variational autoencoders via hypernetworks and trajectory forecasting for autonomous driving.
\newblock {\em arXiv preprint arXiv:2201.09874}, 2022.

\bibitem{park2008multiple}
Youngmin Park, Vincent Lepetit, and Woontack Woo.
\newblock Multiple 3d object tracking for augmented reality.
\newblock In {\em 2008 7th IEEE/ACM International Symposium on Mixed and Augmented Reality}, pages 117--120. IEEE, 2008.

\bibitem{qi2019amodal}
Lu~Qi, Li~Jiang, Shu Liu, Xiaoyong Shen, and Jiaya Jia.
\newblock Amodal instance segmentation with kins dataset.
\newblock In {\em Proceedings of the IEEE/CVF Conference on Computer Vision and Pattern Recognition}, pages 3014--3023, 2019.

\bibitem{qian2023impdet}
Xuelin Qian, Li~Wang, Yi~Zhu, Li~Zhang, Yanwei Fu, and Xiangyang Xue.
\newblock Impdet: Exploring implicit fields for 3d object detection.
\newblock In {\em Proceedings of the IEEE/CVF Winter Conference on Applications of Computer Vision}, pages 4260--4270, 2023.

\bibitem{sun2022amodal}
Yihong Sun and Adam Kortylewski.
\newblock Amodal segmentation through out-of-task and out-of-distribution generalization with a bayesian model. cvpr. 2022.
\newblock In {\em IEEE Conference on Computer Vision and Pattern Recognition}, 2022.

\bibitem{tangemann2021unsupervised}
Matthias Tangemann, Steffen Schneider, Julius von K{\"u}gelgen, Francesco Locatello, Peter Gehler, Thomas Brox, Matthias K{\"u}mmerer, Matthias Bethge, and Bernhard Sch{\"o}lkopf.
\newblock Unsupervised object learning via common fate.
\newblock {\em arXiv preprint arXiv:2110.06562}, 2021.

\bibitem{tran2022aisformer}
Minh Tran, Khoa Vo, Kashu Yamazaki, Arthur Fernandes, Michael Kidd, and Ngan Le.
\newblock Aisformer: Amodal instance segmentation with transformer.
\newblock {\em arXiv preprint arXiv:2210.06323}, 2022.

\bibitem{vaswani2017attention}
Ashish Vaswani, Noam Shazeer, Niki Parmar, Jakob Uszkoreit, Llion Jones, Aidan~N Gomez, {\L}ukasz Kaiser, and Illia Polosukhin.
\newblock Attention is all you need.
\newblock {\em Advances in neural information processing systems}, 30, 2017.

\bibitem{xiao2021amodal}
Yuting Xiao, Yanyu Xu, Ziming Zhong, Weixin Luo, Jiawei Li, and Shenghua Gao.
\newblock Amodal segmentation based on visible region segmentation and shape prior.
\newblock In {\em Proceedings of the AAAI Conference on Artificial Intelligence}, volume~35, pages 2995--3003, 2021.

\bibitem{yang2019embodied}
Jianwei Yang, Zhile Ren, Mingze Xu, Xinlei Chen, David~J Crandall, Devi Parikh, and Dhruv Batra.
\newblock Embodied amodal recognition: Learning to move to perceive objects.
\newblock In {\em Proceedings of the IEEE/CVF International Conference on Computer Vision}, pages 2040--2050, 2019.

\bibitem{yao2022self}
Jian Yao, Yuxin Hong, Chiyu Wang, Tianjun Xiao, Tong He, Francesco Locatello, David Wipf, Yanwei Fu, and Zheng Zhang.
\newblock Self-supervised amodal video object segmentation.
\newblock {\em arXiv preprint arXiv:2210.12733}, 2022.

\bibitem{zhan2020self}
Xiaohang Zhan, Xingang Pan, Bo~Dai, Ziwei Liu, Dahua Lin, and Chen~Change Loy.
\newblock Self-supervised scene de-occlusion.
\newblock In {\em Proceedings of the IEEE/CVF Conference on Computer Vision and Pattern Recognition}, pages 3784--3792, 2020.

\bibitem{zhao2023fast}
Xu~Zhao, Wenchao Ding, Yongqi An, Yinglong Du, Tao Yu, Min Li, Ming Tang, and Jinqiao Wang.
\newblock Fast segment anything.
\newblock {\em arXiv preprint arXiv:2306.12156}, 2023.

\bibitem{zhu2017semantic}
Yan Zhu, Yuandong Tian, Dimitris Metaxas, and Piotr Doll{\'a}r.
\newblock Semantic amodal segmentation.
\newblock In {\em Proceedings of the IEEE conference on computer vision and pattern recognition}, pages 1464--1472, 2017.

\end{thebibliography}
}


\clearpage

\appendix

\counterwithin{figure}{section}
\counterwithin{table}{section}

\section{Appendix / supplemental material}



\subsection{Broader Impacts}
\label{BroaderImpacts}

\textbf{1) Positive Societal Impacts}
Our model can accurately predict the shapes of obscured cars or pedestrians in autonomous driving scenarios, potentially preventing accidents. Additionally, our approach can be applied to robotic grasping, helping robots better locate and handle objects they need to grasp.

\textbf{2) Negative Societal Impacts}
Training on the KINS and COCOA-cls datasets to complete masks of cars or pedestrians might raise privacy concerns.

\subsection{Additional Visualizations}
\begin{figure}[h]
\begin{center}
\includegraphics[width=0.95\columnwidth]{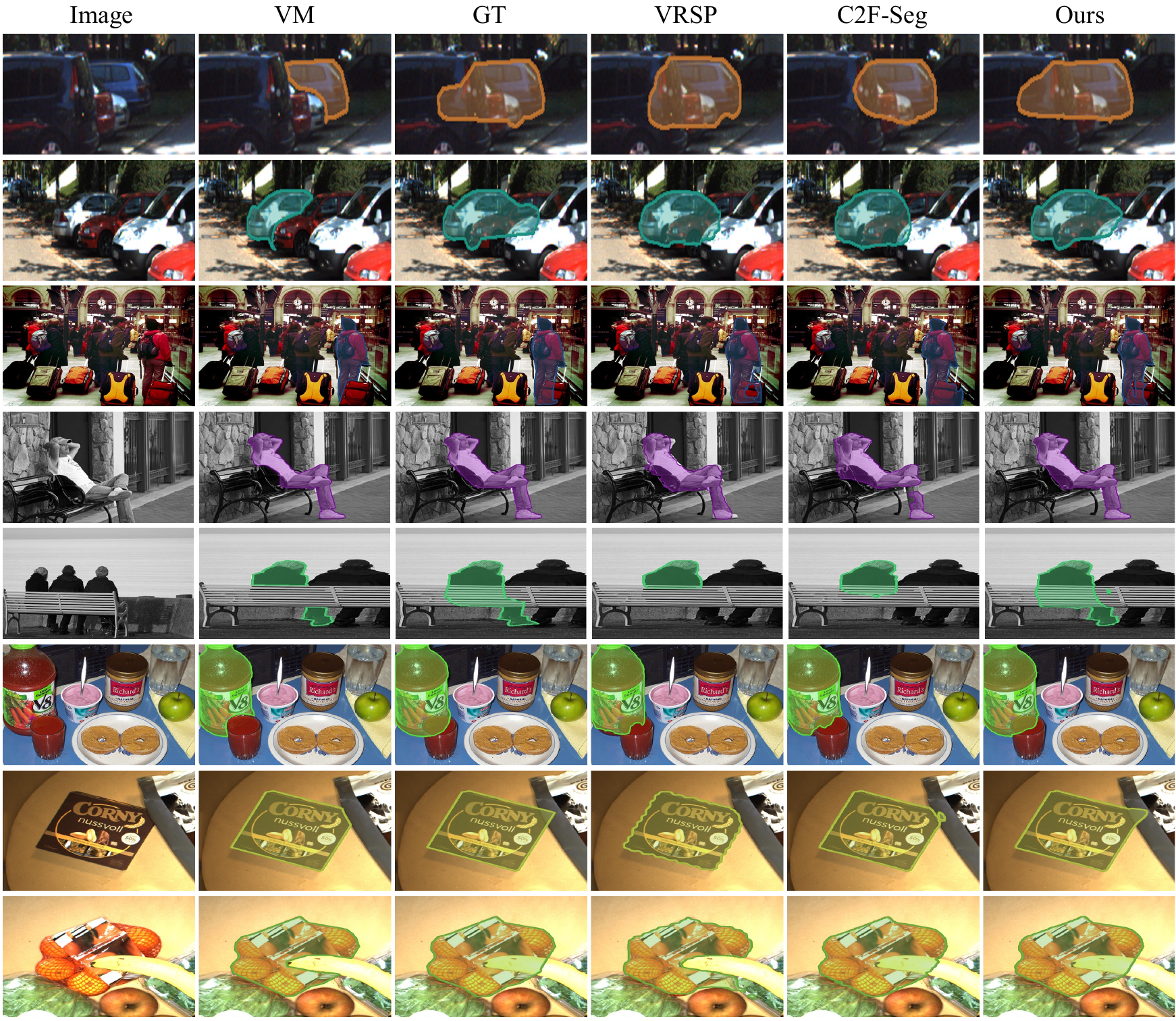}
\vskip -0.05in
\caption{More Qualitative results compared with VRSP, C2F-Seg, and our model.}
\label{fig:visualization2}
\end{center}
\vskip -0.1in
\end{figure}

\begin{figure}
\begin{center}
\includegraphics[width=0.95\columnwidth]{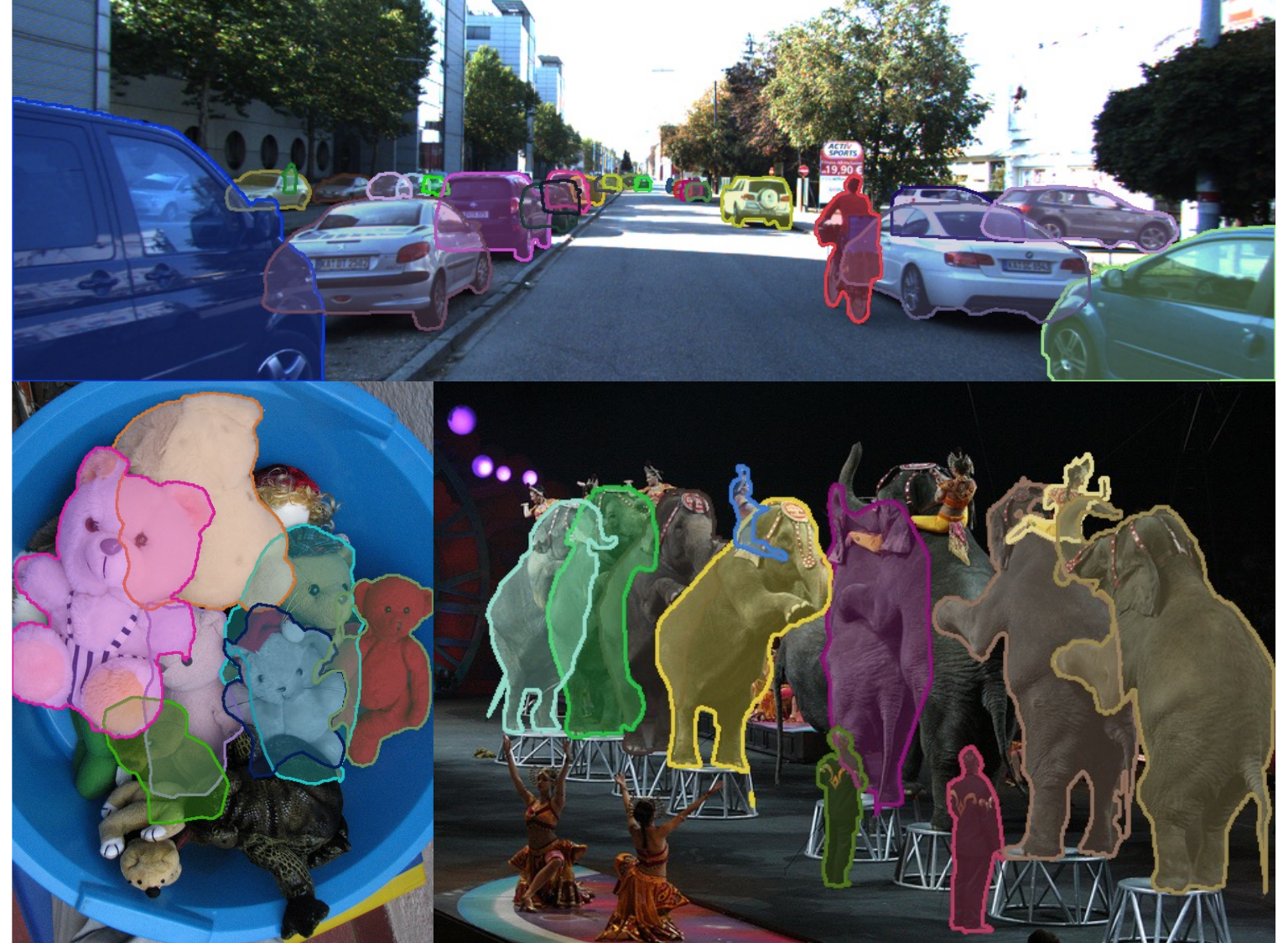}
\vskip -0.05in
\caption{More Visualization of Amodal masks on KINS and COCOA-cls.}
\label{fig:full_figure}
\end{center}
\vskip -0.1in
\end{figure}

To better showcase the performance of our model, we provide additional visualization results. As shown in Fig.~\ref{fig:visualization2}, we present further comparisons between our model and the baseline. Fig.~\ref{fig:full_figure} demonstrates our model's capability to complete the masks of all objects in an image (Note: We complete the objects with the visible mask annotations provided in the dataset).

\end{document}